\documentclass{article} 
\usepackage{iclr2026_conference,times}

\usepackage{amsmath,amsfonts,bm}

\def\eqref#1{equation~\ref{#1}}

\def\1{\bm{1}}

\DeclareMathAlphabet{\mathsfit}{\encodingdefault}{\sfdefault}{m}{sl}
\SetMathAlphabet{\mathsfit}{bold}{\encodingdefault}{\sfdefault}{bx}{n}

\usepackage{hyperref}
\usepackage{url}
\usepackage{tcolorbox}
\usepackage{listings}
\usepackage{amssymb}
\usepackage{booktabs}
\usepackage{graphicx}
\usepackage{longtable}
\usepackage{booktabs}
\usepackage{tabularx}
\usepackage{ragged2e}
\usepackage{array}
\usepackage{microtype}

\newcolumntype{Y}{>{\RaggedRight\arraybackslash}X}

\title{Navigating the Concept Space of Language Models}

\author{Wilson E. Marcílio-Jr \thanks{wilson.marcilio@unesp.br.} \\
Adaption Labs\\
Curitiba, PR, Brazil \\
\texttt{wilson@adaptionlabs.ai}
\And
Danilo Medeiros Eler \\
São Paulo State University (UNESP)\\
Presidente Prudente, SP, Brazil \\ 
\texttt{danilo.eler@unesp.br}\\
}

\iclrfinalcopy 
\begin{document}

\maketitle

\begin{abstract}
Sparse autoencoders (SAEs) trained on large language model activations output thousands of features that enable mapping to human-interpretable concepts. The current practice for analyzing these features primarily relies on inspecting top-activating examples, manually browsing individual features, or performing semantic search on interested concepts, which makes exploratory discovery of concepts difficult at scale. In this paper, we present \emph{Concept Explorer}, a scalable interactive system for post-hoc exploration of SAE features that organizes concept explanations using hierarchical neighborhood embeddings. Our approach constructs a multi-resolution manifold over SAE feature embeddings and enables progressive navigation from coarse concept clusters to fine-grained neighborhoods, supporting discovery, comparison, and relationship analysis among concepts. We demonstrate the utility of Concept Explorer on SAE features extracted from SmolLM2, where it reveals coherent high-level structure, meaningful subclusters, and distinctive rare concepts that are hard to identify with existing workflows.
\end{abstract}

\section{Introduction}

Sparse autoencoders (SAEs) trained on internal activations of large language models have recently emerged as an effective approach for discovering latent features that align with human-interpretable concepts~\citep{bricken2023monosemanticity, huben2024sparse, ONeill2024DisentanglingDE, Paulo2024AutomaticallyIM}. By enforcing sparsity in a bottleneck representation, SAEs encourage a decomposition of activation space into a large set of relatively independent feature directions, many of which correspond to syntactic, semantic, or functional patterns. This paradigm has enabled mechanistic interpretability studies that operate at a scale beyond individual neurons.

At the same time, advances in SAE training have substantially increased the number of learned features. Contemporary SAEs frequently contain hundreds of thousands to millions of latent dimensions~\citep{templeton2024scaling, gao2025scaling}. While this scale improves representational coverage, it introduces a fundamental analysis challenge: only a small fraction of features can be examined manually, yet there is currently no principled mechanism for organizing or surveying the full feature set. Consequently, the dominant bottleneck has shifted from feature learning to feature discovery.

Common workflows involve selecting individual features, retrieving their top-activating contexts, using an LLM to infer their semantics, and employ heuristics to identify interesting concepts~\citep{Cunningham2023SparseAF, Paulo2024AutomaticallyIM}. Such heuristics include ranking features by activation frequency, selectivity, or magnitude, and performing similarity search around known features~\citep{zhou2025llm}. Although effective for validating specific hypotheses, these strategies are inherently local and query-driven. They provide limited support for global reasoning over the feature space and do not naturally facilitate open-ended discovery of concept families, identification of rare or isolated features, or analysis of relationships among large groups of features. On the other hand, single-level interaction embeddings like neuropedia.org or the one utilized by~\citet{Cunningham2023SparseAF} make feature discovery difficult due to visualization clutter.

Given a large collection of SAE features together and their textual explanations, an effective visual representation should support coarse-to-fine exploration, preserve neighborhood relationships among related features, and scale to millions of items. This representation should provide both a global overview and local inspection through interactive analysis.

We introduce \emph{Concept Explorer}\footnote{Code and application will be released after double-blind revision.}, a scalable interactive system for post-hoc exploration of SAE features based on hierarchical neighborhood embeddings. Feature explanations are embedded into a multi-resolution manifold constructed using HUMAP~\citep{marciliojr_humap2024}, yielding a hierarchy of increasingly coarse representations. This hierarchy enables progressive navigation from high-level concept groupings to fine-grained neighborhoods, while preserving both local and global structure. Concept Explorer supports identification of dominant concept families, discovery of rare or disconnected features, and examination of relationships among concepts.

Our work does not propose a new SAE training objective or architecture. Instead, we focus on the downstream problem of organizing and exploring the outputs of existing SAE methods.

\section{Background}

\subsection{Sparse Autoencoders for Large Language Model Representations}

Sparse autoencoders (SAEs) provide a scalable method for decomposing high-dimensional activation vectors of large language models (LLMs) into sparse, interpretable latent features. Given an activation vector $h \in \mathbb{R}^D$ from a fixed LLM layer, an SAE learns a sparse code $
z \in \mathbb{R}^K, K \gg D$, and a reconstruction $\hat{h}$ such that $h \approx \hat{h}$. Then, an encoder--decoder parameterization is used:
\begin{align}
z &= f(W_e h + b_e), \\
\hat{h} &= g(W_d z + b_d),
\end{align}
where $W_e \in \mathbb{R}^{K \times D}$, $W_d \in \mathbb{R}^{D \times K}$, and $f$ and $g$ are often the same activation functions. The model minimizes the reconstruction loss with a sparsity constraint, which ensures that only a small subset of features contribute to each reconstruction.

SAEs are trained on large corpora of LLM activations with the base model frozen. The learned features can be interpreted by identifying inputs that maximize individual latent activations or by performing causal interventions using the decoder directions.

\section{HUMAP for hierarchical neighborhood embedding}

HUMAP constructs a hierarchy of subsets based on landmarks and the corresponding fuzzy similarity graphs, enabling visualization of multiple resolution manifolds through repeated application of UMAP~\citep{McInnes2018UMAPUM}. 

Given  $X^{(0)} = \{x_i\}_{i=1}^n \in \mathbb{R}^D$ and a nested sequence $X^{(0)} \supset X^{(1)} \supset \cdots \supset X^{(L)}$, HUMAP computes an embedding $Y^{(\ell)} = \{y_i^{(\ell)}\} \in \mathbb{R}^2$ at each level $\ell$. Then, for each $x_i, x_j \in X^{(\ell)}$, it defines connection strength weights ($p_{ij}$) using the UMAP kernel. Finally, these weights are then used to induce a Markov transition matrix
\begin{equation}
T_{ij} =
\frac{p_{ij}}{\sum_{m \in N_k(i)} p_{im}}.
\end{equation}

Landmarks are selected by ranking nodes according to visit frequency after random walks on $T$, which generates level $X^{(\ell+1)}$. For each landmark $u \in $, a representation neighborhood is encoded in a sparse matrix $R$ and lardmark similarities are computed using Equation~\ref{eq:s}, and each level embedding $Y^{(\ell)}$ is obtained by applying UMAP to $S$.

\begin{equation}
\label{eq:s}
S = 1 - \frac{R^\top R}{max(R^\top R)}.
\end{equation}

Finally, random walks create a region of influence at level $X^{(\ell)}$ for each landmark in level $X^{(\ell+1)}$, enabling hierarchical drill-down. That is, one can project on $X^{(\ell)}$ the landmarks of interest in $X^{(\ell+1)}$, essentially projecting the region of influence (more data) from the selected points.

\section{Methods}

In order to provide scalable and interactive concept finding and analysis we built \emph{Concept Explorer} on top of a hierarchical neighborhood embedding computed using HUMAP. In the following, we explain the design and provide a Use Case in Section~\ref{sec:use-case}.

\subsection{Concept Explorer}
\label{sec:concept-explorer}

We defined the following questions to drive the visualization and interaction design:

\begin{itemize}
    \item \textbf{Q1}: What are the main concepts in the dataset?
    \item \textbf{Q2}: What are the rarest concepts?
    \item \textbf{Q3}: How do concepts interact (overlap and influence)?
\end{itemize}

\begin{figure}[h]
\begin{center}
\includegraphics[width=0.8\linewidth]{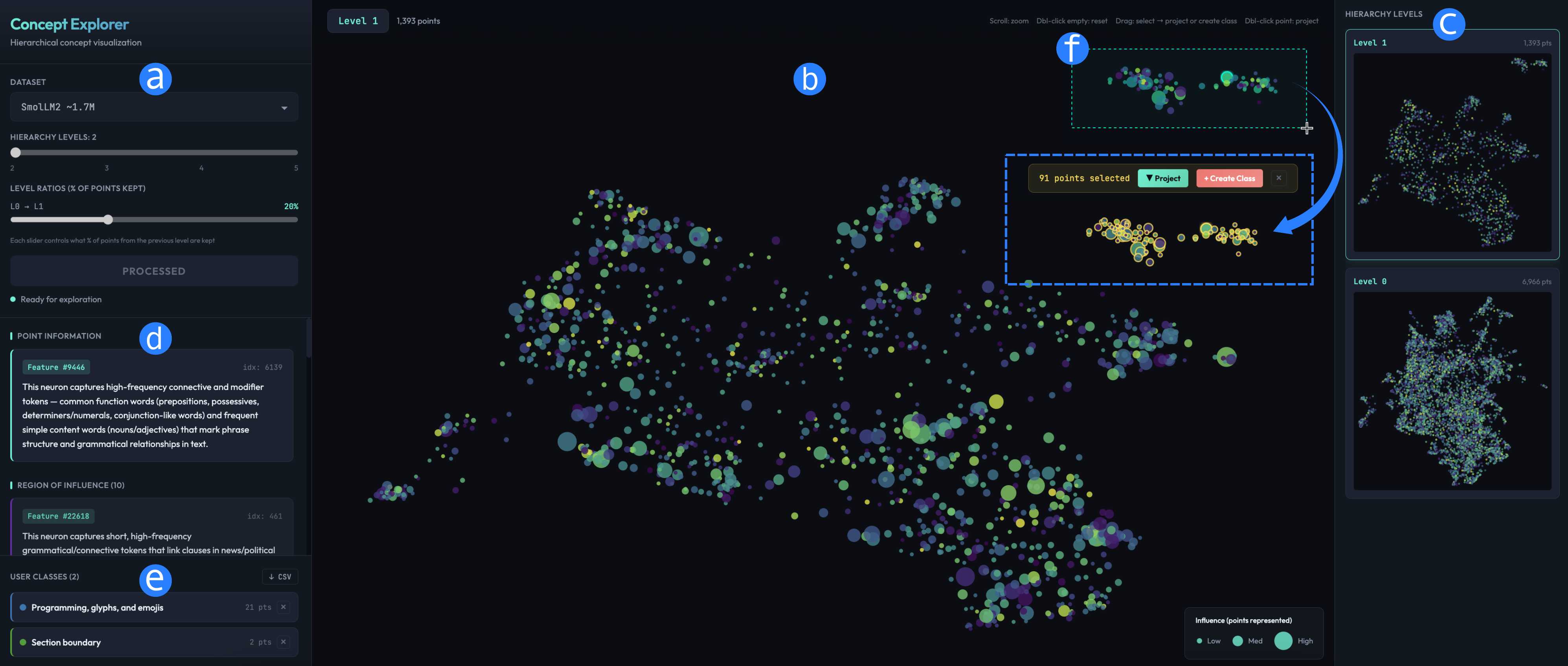}
\end{center}
\caption{Concept Explorer user interface. The right panel shows HUMAP projections at two levels; the left panel shows explanation content, top-k activation contexts and annotation controls for a selected feature. The middle panel shows the projection level being explored.}
\label{fig:tool}
\end{figure}

Figure~\ref{fig:tool} gives an overview of the interface: a multi-level canvas showing HUMAP projections at each level (\textbf{b}, \textbf{c}), a detail panel containing explanation text and top activation contexts for the selected feature (\textbf{d}), and controls for annotating, merging or reprojecting regions of interest (\textbf{e}, \textbf{f}).

To answer \textbf{Q1}, users build a hierarchy with an adjustable number of landmarks per level (the more landmarks the finer is the granularity). Since HUMAP preserves locality and influence regions, similar features remain spatially close across adjacent levels; users can therefore iterate by building more levels and by inspecting the textual explanations and top-k contexts for each feature.

To answer \textbf{Q2}, Concept Explorer exposes three lightweight signals useful for triage: (i) projection proximity (points far from any dense group are candidate outliers), (ii) influence-region size (small regions often indicate specialized, rare features), and (iii) explanation similarity (text embeddings of explanations allow rapid filtering for duplicated or highly-overlapping concepts).

To answer \textbf{Q3}, HUMAP's hierarchical layout maintains both local (intra cluster) and global (among clusters) structure, and the interface lets users reproject the region-of-influence of a landmark on the level below while preserving manifold relations (see Figure~\ref{fig:tool}\textbf{F}).

\section{Use Case - Exploring Concepts from SmolLM2}
\label{sec:use-case}

In this use case we demonstrate Concept Explorer on features extracted by a SAE trained on SmolLM2's~\citep{allal2025smollm2smolgoesbig} MLPs by Eleuther AI~\citep{EleutherAI_sae_SmolLM2_135M_64x_2025}. We processed 1.7M sentences and collected the top-16 activation contexts per SAE feature from a chosen bottleneck layer (layer 27 in our experiments). For each feature we generated a short textual explanation using GPT5-mini; the prompt template is shown below.

\begin{tcolorbox}
The following is a list of contexts where the neuron activates for the token in $<<$ $>>$.
Taking the tokens in $<<$token$>>$ into account and the list of contexts, please provide a short description that captures the meaning of the neuron.

Start with your explanation as "This neuron captures...". Be concise but not too vague.
\end{tcolorbox}

We embedded the explanations using \texttt{nomic-embed-text-v1.5}~\citep{nussbaum2025nomicembedtrainingreproducible} for the 36864 features and stored their metadata: (i) the explanation text, (ii) its embedding, (iii) top-k contexts, and (iv) the SAE feature id. These metadata are the only inputs required by Concept Explorer: the interface uses the explanation embeddings to group semantically similar explanations and HUMAP to visualize the activation latent structure.

\begin{figure}[h]
\begin{center}
\includegraphics[width=0.7\linewidth]{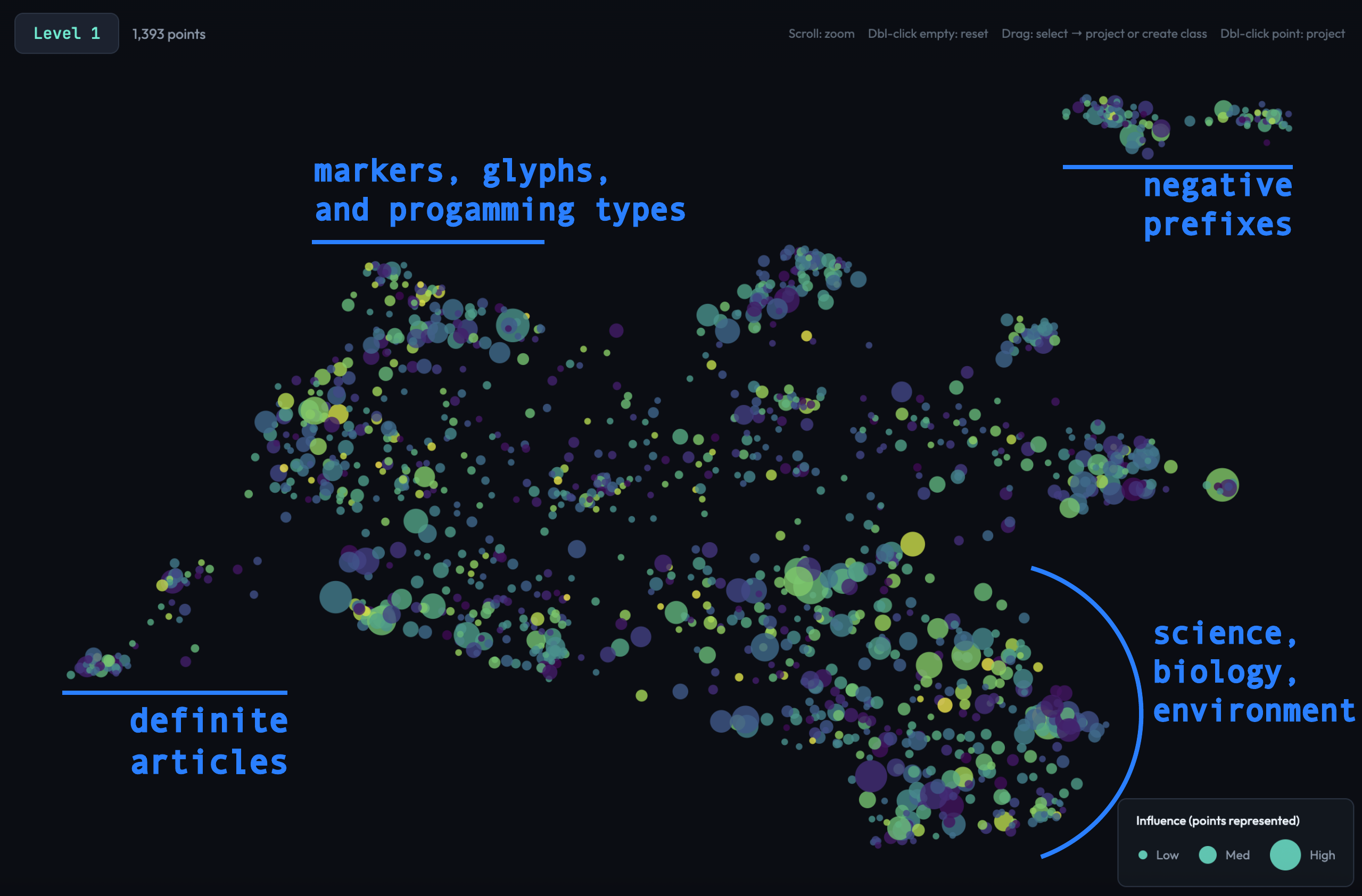}
\end{center}
\caption{HUMAP projection of explanation embeddings for SAE features from SmolLM2 (1.5M contexts, top-16 per feature). Colors indicate analyst-assigned coarse categories.}
\label{fig:use-case-plot}
\end{figure}

Figure~\ref{fig:use-case-plot} shows the HUMAP projection for the explanations. The projection reveals multiple disconnected clusters with a mixture of broad influence regions (many explanations overlap in embedding space) and several tightly localized, low-influence features (rare or highly specific)---these are easily identifiable through circle area. The projection already helps on answering questions like \textbf{Q1}. Below we summarize a few notable findings.

\subsection{Punctuation cluster}

\begin{figure}[h]
\begin{center}
\includegraphics[width=0.7\linewidth]{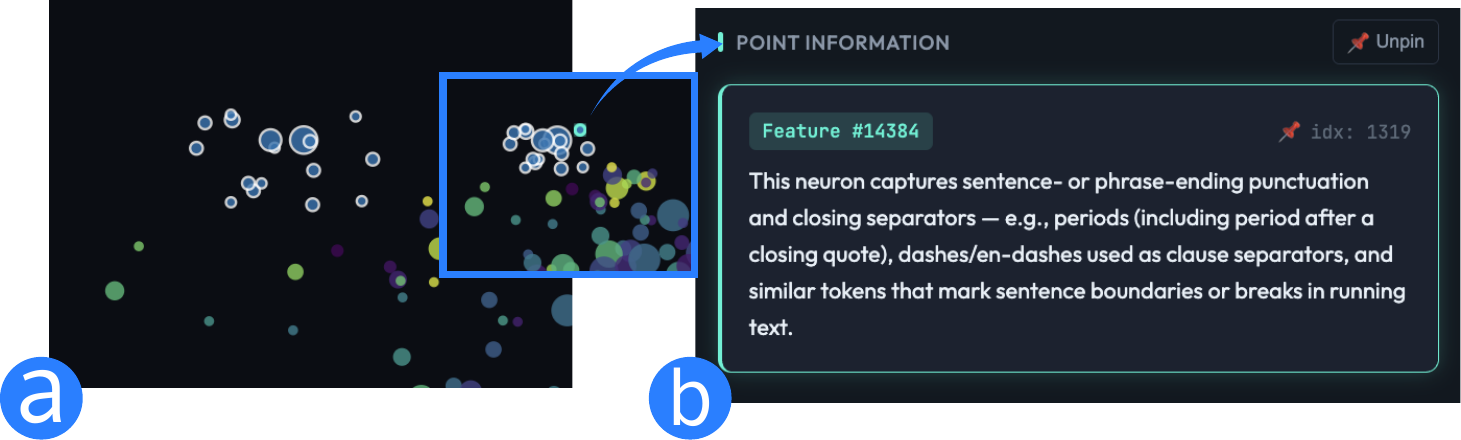}
\end{center}
\caption{(a) Punctuation-related features clustered by explanation embeddings; (b) a rare feature focused on dashes and list markers. Table~\ref{tab:feature-ids-explanations} shows relevant explanations for feature ids.}
\label{fig:punctuation-cluster}
\end{figure}

Figure~\ref{fig:punctuation-cluster} shows how the explanations are projected onto the second hierarchical level. We can see diversity on the feature explanations---also highlighted by the manual annotations. For instance, on the top-left, feature \#12378 activates on sentence-end patterns (periods, ellipses) while feature \#22970 fires for a range of sentence-boundary markers.

By focusing on a small influence region, we discovered features that specialize on dashes / em-dashes and on list-boundary markers (patterns that are frequent enough to be distributed across the dataset but are often missed by global clustering methods due to sparsity of contexts). Such a cluster is highlighted in Figure~\ref{fig:punctuation-cluster}(\textbf{a}), while (\textbf{b}) shows a single rare feature discovered inside that cluster.

\begin{figure}[h]
\begin{center}
\includegraphics[width=0.7\linewidth]{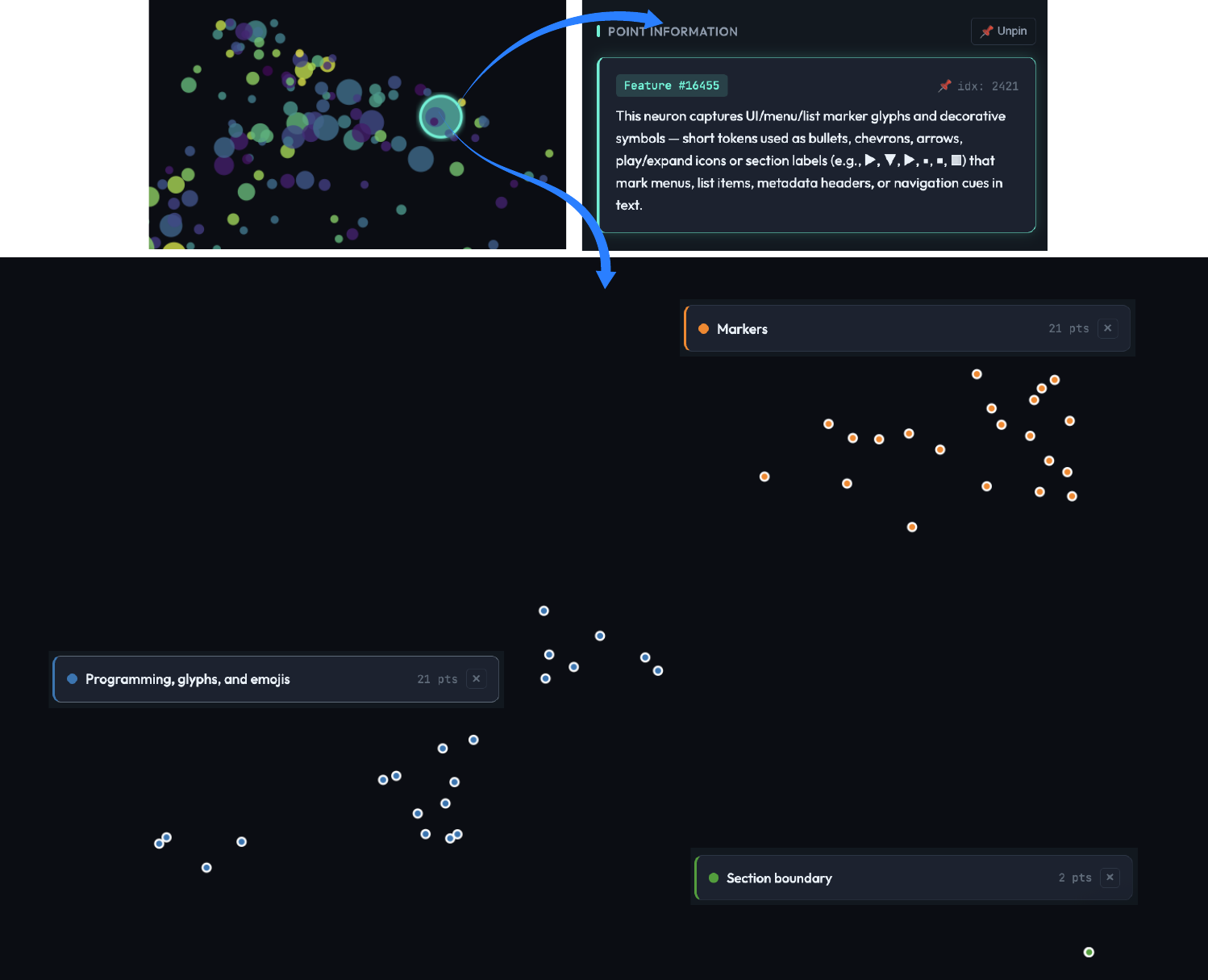}
\end{center}
\caption{Marker and glyph concepts discovered by Concept Explorer. Reprojection of a level-1 region of influence to level 0 reveals fine-grained substructure.}
\label{fig:marker-glyphs}
\end{figure}

Figure~\ref{fig:marker-glyphs} shows another example concerning marker/glyph features, where one landmark at level 1 has 20 features in its region of influence. Reprojecting that region to level 0 exposes subclusters and emphasizes rare concepts and the relationship among clusters (\textbf{Q2} and \textbf{Q3}): programming tokens and glyphs (left), placeholder markers like ``$<<>>$'' and Java-style annotations such as ``@Override''(top-right), and enumeration/list separators (bottom-right).

\section{Conclusion}

We addressed the problem of scalable post-hoc exploration of sparse autoencoder features, which arise in large numbers in modern interpretability pipelines and are difficult to analyze with existing feature-centric workflows. We introduced Concept Explorer, a system that organizes feature explanations using hierarchical neighborhood embeddings, enabling multi-resolution navigation of large concept spaces while preserving local and global structure.

Through a case study on SAE features derived from SmolLM2, we showed that this representation supports identification of coherent high-level concept families, analysis of internal substructure, and discovery of rare or isolated features, suggesting that hierarchical manifold-based organization is a useful abstraction for managing large collections of learned concepts.. These capabilities facilitate open-ended concept discovery that is not well supported by current SAE analysis practices.

\bibliography{iclr2026_conference}

@inproceedings{
huben2024sparse,
title={Sparse Autoencoders Find Highly Interpretable Features in Language Models},
author={Robert Huben and Hoagy Cunningham and Logan Riggs Smith and Aidan Ewart and Lee Sharkey},
booktitle={The Twelfth International Conference on Learning Representations},
year={2024},
url={https://openreview.net/forum?id=F76bwRSLeK}
}

@article{ONeill2024DisentanglingDE,
  title={Disentangling Dense Embeddings with Sparse Autoencoders},
  author={Charles O'Neill and Christine Ye and Kartheik G. Iyer and John F. Wu},
  journal={ArXiv},
  year={2024},
  volume={abs/2408.00657},
  url={https://api.semanticscholar.org/CorpusID:271601116}
}

@article{Paulo2024AutomaticallyIM,
  title={Automatically Interpreting Millions of Features in Large Language Models},
  author={Gonccalo Paulo and Alex Troy Mallen and Caden Juang and Nora Belrose},
  journal={ArXiv},
  year={2024},
  volume={abs/2410.13928},
  url={https://api.semanticscholar.org/CorpusID:273482460}
}

@misc{nussbaum2025nomicembedtrainingreproducible,
      title={Nomic Embed: Training a Reproducible Long Context Text Embedder}, 
      author={Zach Nussbaum and John X. Morris and Brandon Duderstadt and Andriy Mulyar},
      year={2025},
      eprint={2402.01613},
      archivePrefix={arXiv},
      primaryClass={cs.CL},
      url={https://arxiv.org/abs/2402.01613}, 
}

@inproceedings{
zhou2025llm,
title={{LLM} Neurosurgeon: Targeted Knowledge Removal in {LLM}s using Sparse Autoencoders},
author={Dylan Zhou and Kunal Patil and Yifan Sun and Karthik lakshmanan and Senthooran Rajamanoharan and Arthur Conmy},
booktitle={ICLR 2025 Workshop on Building Trust in Language Models and Applications},
year={2025},
url={https://openreview.net/forum?id=aeQeXlG2Pw}
}

@article{templeton2024scaling,
       title={Scaling Monosemanticity: Extracting Interpretable Features from Claude 3 Sonnet},
       author={Templeton, Adly and Conerly, Tom and Marcus, Jonathan and Lindsey, Jack and Bricken, Trenton and Chen, Brian and Pearce, Adam and Citro, Craig and Ameisen, Emmanuel and Jones, Andy and Cunningham, Hoagy and Turner, Nicholas L and McDougall, Callum and MacDiarmid, Monte and Freeman, C. Daniel and Sumers, Theodore R. and Rees, Edward and Batson, Joshua and Jermyn, Adam and Carter, Shan and Olah, Chris and Henighan, Tom},
       year={2024},
       journal={Transformer Circuits Thread},
       url={https://transformer-circuits.pub/2024/scaling-monosemanticity/index.html}
    }

@article{bricken2023monosemanticity,
       title={Towards Monosemanticity: Decomposing Language Models With Dictionary Learning},
       author={Bricken, Trenton and Templeton, Adly and Batson, Joshua and Chen, Brian and Jermyn, Adam and Conerly, Tom and Turner, Nick and Anil, Cem and Denison, Carson and Askell, Amanda and Lasenby, Robert and Wu, Yifan and Kravec, Shauna and Schiefer, Nicholas and Maxwell, Tim and Joseph, Nicholas and Hatfield-Dodds, Zac and Tamkin, Alex and Nguyen, Karina and McLean, Brayden and Burke, Josiah E and Hume, Tristan and Carter, Shan and Henighan, Tom and Olah, Christopher},
       year={2023},
       journal={Transformer Circuits Thread},
       note={https://transformer-circuits.pub/2023/monosemantic-features/index.html}
    }

@inproceedings{
gao2025scaling,
title={Scaling and evaluating sparse autoencoders},
author={Leo Gao and Tom Dupre la Tour and Henk Tillman and Gabriel Goh and Rajan Troll and Alec Radford and Ilya Sutskever and Jan Leike and Jeffrey Wu},
booktitle={The Thirteenth International Conference on Learning Representations},
year={2025},
url={https://openreview.net/forum?id=tcsZt9ZNKD}
}

@article{Cunningham2023SparseAF,
  title={Sparse Autoencoders Find Highly Interpretable Features in Language Models},
  author={Hoagy Cunningham and Aidan Ewart and Logan Riggs Smith and Robert Huben and Lee Sharkey},
  journal={ArXiv},
  year={2023},
  volume={abs/2309.08600},
  url={https://api.semanticscholar.org/CorpusID:261934663}
}

@ARTICLE{marciliojr_humap2024,
        author={Marcílio-Jr, Wilson E. and Eler, Danilo M. and Paulovich, Fernando V. and Martins, Rafael M.},
        journal={IEEE Transactions on Visualization and Computer Graphics},
        title={HUMAP: Hierarchical Uniform Manifold Approximation and Projection},
        year={2024},
        volume={},
        number={},
        pages={1-10},
        doi={10.1109/TVCG.2024.3471181}
}

@article{McInnes2018UMAPUM,
  title={UMAP: Uniform Manifold Approximation and Projection for Dimension Reduction},
  author={Leland McInnes and John Healy},
  journal={ArXiv},
  year={2018},
  volume={abs/1802.03426},
  url={https://api.semanticscholar.org/CorpusID:3641284}
}

@misc{allal2025smollm2smolgoesbig,
      title={SmolLM2: When Smol Goes Big -- Data-Centric Training of a Small Language Model}, 
      author={Loubna Ben Allal and Anton Lozhkov and Elie Bakouch and Gabriel Martín Blázquez and Guilherme Penedo and Lewis Tunstall and Andrés Marafioti and Hynek Kydlíček and Agustín Piqueres Lajarín and Vaibhav Srivastav and Joshua Lochner and Caleb Fahlgren and Xuan-Son Nguyen and Clémentine Fourrier and Ben Burtenshaw and Hugo Larcher and Haojun Zhao and Cyril Zakka and Mathieu Morlon and Colin Raffel and Leandro von Werra and Thomas Wolf},
      year={2025},
      eprint={2502.02737},
      archivePrefix={arXiv},
      primaryClass={cs.CL},
      url={https://arxiv.org/abs/2502.02737}, 
}

@online{EleutherAI_sae_SmolLM2_135M_64x_2025,
  title        = {EleutherAI/sae-SmolLM2-135M-64x},
  author       = {{EleutherAI}},
  year         = {2025},
  url          = {https://huggingface.co/EleutherAI/sae-SmolLM2-135M-64x},
  note         = {Accessed: 2026-01-25},
  organization = {Hugging Face}
}
\bibliographystyle{iclr2026_conference}

\appendix
\section{Appendix}
\label{sec:appendix}

\begin{longtable}{p{2cm} p{1cm} p{10cm}}
\toprule
Class Name                      & Feature Id & Explanation  (\textit{This neuron captures...})                                                                                                                                                                                                                                                                                                                                                                                                                                                                                                                                                                        \\* \midrule
\endfirsthead
\endhead
\bottomrule
\endfoot
\endlastfoot
Programming, glyphs, and emojis & 9529       &  short interface/command tokens highlighted with double-angle-bracket markup — i.e., keyboard keys, button or menu labels, quoted literal strings, and brief instruction verbs/affixes (like "click", "Start", "Esc", or "un" in "unplug") used in step-by-step UI/instruction text.                                                                                                                                                                                                                                                                            \\
Programming, glyphs, and emojis & 6967       &  short inline label/prefix tokens used as decorative/attention badges at the start of headlines or UI items—e.g.,  emojis, arrows or symbols placed in \textless{}\textless{}...\textgreater{}\textgreater to mark categories, new items, or section headings.                                                                                                                                                                                                                                        \\
Programming, glyphs, and emojis & 18379      &  imperative UI-action tokens—especially the verb "click"—used to mark step-by-step instructions in software/docs (often highlighted inside \textless{}\textless \textgreater{}\textgreater{}).                                                                                                                                                                                                                                                                                                                                                                  \\
Programming, glyphs, and emojis & 5466       &  short tokens used in instructional/UI contexts — especially single-letter key names and function-key labels (e.g., C, V, F, R), lone punctuation used in shortcut/list formatting (comma, colon, semicolon, arrow), and brief connective words that appear in commands or how‑to steps (like, for, until, become, eventually).                                                                                                                                                                                                                                 \\
Programming, glyphs, and emojis & 11714      &  short, standalone identifier-like tokens — e.g., file/config extensions and CLI keywords (yml, yaml, compile), brief abbreviations/initialisms (M in M.V.P., MX), and compact content words (eyes, dry, nasal). In other words, it responds to short label/identifier tokens that commonly appear in code/command contexts and short noun/adjective tokens in prose.                                                                                                                                                                                           \\
Programming, glyphs, and emojis & 16455      &  UI/menu/list marker glyphs and decorative symbols — short tokens used as bullets, chevrons, arrows, play/expand icons or section labels that mark menus, list items, metadata headers, or navigation cues in text.                                                                                                                                                                                                                                                                                                                    \\
Programming, glyphs, and emojis & 13277      &  short, content-bearing tokens used as labels, commands, or identifiers — especially single-letter/initial tokens (like the keybind or class initial "C") and short noun/verb fragments or contraction pieces that appear as standalone tokens in code, UI, instructions, and game/control contexts.                                                                                                                                                                                                                                                            \\
Programming, glyphs, and emojis & 18121      &  tokens associated with front-end UI and React/JSX code contexts — e.g., component/tag names (Header, Footer, App, Error, SearchBar), UI-related words (clean, customizable), and common JS/React syntax like the arrow callback (=\textgreater{}) used in map/filter operations.                                                                                                                                                                                                                                                                               \\
Programming, glyphs, and emojis & 598        &  decorative dingbat/marker characters used as bullets, menu/section markers, or “new” labels that precede headings, list items, buttons, or UI prompts — i.e., typographic/ornamental markers rather than substantive words.                                                                                                                                                                                                                                                                                                           \\
Programming, glyphs, and emojis & 5954       &  small UI/formatting glyphs and decorative non-word icons used as section/menu/list markers (e.g., arrows, bullets, square labels like "metadata" ), typically appearing before headings, menu items, or navigation controls.                                                                                                                                                                                                                                                                                                                  \\
Programming, glyphs, and emojis & 3522       &  references to keyboard keys and shortcut notations — especially modifier and command keys (e.g., Ctrl, Alt, Shift, Win/Windows, Delete, R, X, Cmd) used in instructional/step-by-step text.                                                                                                                                                                                                                                                                                                                                                                    \\
Programming, glyphs, and emojis & 9754       &  declaration- and type-related tokens in C/C++/GLSL graphics code — e.g., keywords like "struct", type names like "string", and identifiers/markers for vertex/texture/uniform fields (vertex, UV/TexCoords, MeshData, getTextureCoords, etc.).                                                                                                                                                                                                                                                                                                                 \\
Programming, glyphs, and emojis & 18302      &  short, non-syntactic label-like tokens and morpheme fragments — i.e., UI/placeholders and standalone bits such as "Image" placeholders, short words embedded in compounds (e.g., "stop" in Showstopper), two-letter uppercase tags (e.g., "BN"), or isolated characters.                                                                                                                                                                                                                                                                                       \\
Programming, glyphs, and emojis & 3083       &  small UI/list/disclosure glyphs and decorative markers—icons like arrows/triangles, bullets, boxes, and shading symbols used to mark menus, headings, list items, or other interface/formatting affordances.                                                                                                                                                                                                                                                                                                                               \\
Programming, glyphs, and emojis & 15569      &  tokens that label or annotate graphical or schematic elements—especially color words (blue, red, green) used in figure/legend descriptions and short markers used in labels (including the hyphen in hyphenated labels like dual‑flush or quad‑core).                                                                                                                                                                                                                                                                                                          \\
Programming, glyphs, and emojis & 6501       &  UI/navigation or section-label markers—decorative glyphs and short header tokens that introduce menus, lists, or metadata (e.g., arrows/triangles, bullets, block characters, and brief header words like "metadata" or single-character labels in other languages).                                                                                                                                                                                                                                                                                 \\
Programming, glyphs, and emojis & 5905       &  decorative UI/list markers and short standalone interface glyphs—single-symbol bullets/arrows/squares and similar short labels used to mark menus, list items, expand/collapse controls, or section headings.                                                                                                                                                                                                                                                                                                                         \\
Programming, glyphs, and emojis & 18202      &  tokens that are user-interface element labels — names of tabs, buttons, menu items, options or feature labels used in instructions/documentation (e.g., Partitions, Traceroute, Maps, Permissions, Color, Type, Size, Styles).                                                                                                                                                                                                                                                                                                                                 \\
Programming, glyphs, and emojis & 23494      &  short technical identifier tokens found in Unix/manpage and POD-style documentation — e.g., command and utility names, shell/module/function identifiers and their subword fragments (things like uptime, sh, ln, ovs-dpctl, module::Intro).                                                                                                                                                                                                                                                                                                                   \\
Programming, glyphs, and emojis & 5094       &  UI/navigation glyphs and interface markers — small symbolic tokens used as play/menu arrows, bullets/boxes, “new” badges or other clickable/listing affordances in web and app content.                                                                                                                                                                                                                                                                                                                             \\
Programming, glyphs, and emojis & 24700      &  mentions of data types and type-related terms in programming/documentation contexts — e.g., primitive and composite type names and descriptors (float, double, char, real, primitive, composite), structured/type constructs (matrix, Matrix, pointer, linear, packed), and similar type vocabulary.                                                                                                                                                                                                                                                           \\
Markers                         & 19671      &  tokens that are highlighted/annotated inside double angle brackets (\textless{}\textless{}...\textgreater{}\textgreater{}) — typically short code/log fragments: single letters, flags/options, small identifiers or units (e.g., 0, d, GET, GB, \_ , Server) marked as the target token in code, CLI output, or error messages.                                                                                                                                                                                                                               \\
Markers                         & 11770      &  the "@Override" method-annotation token — i.e., tokens that introduce/mark overridden method declarations (commonly seen before getCount/getItem/getItemId/getView in Java/Android adapter classes).                                                                                                                                                                                                                                                                                                                                                           \\
Markers                         & 10081      &  tokens that are the returned value in Python return statements — i.e., variable names or expressions appearing immediately after "return" (often at the end of a function), such as local variables, lists, dicts, strings, or other expression identifiers.                                                                                                                                                                                                                                                                                                   \\
Markers                         & 7273       &  the "\textgreater{}\textgreater{}" token — consecutive right-angle brackets commonly seen as closing markers in scraped or garbled text (e.g., after inline citations, broken HTML/XML tags/attributes, or other noisy markup/ellipsis-like artifacts).                                                                                                                                                                                                                                                                                                        \\
Markers                         & 19349      &  tokens that are explicitly marked or quoted with double-angle brackets (\textless{}\textless \textgreater{}\textgreater{}) — i.e., words presented as highlighted/annotated/quoted items (a formatting/markup signal) rather than their specific lexical meaning.                                                                                                                                                                                                                                                                                              \\
Markers                         & 20480      &  tokens that are being marked/highlighted with double angle brackets (\textless{}\textless{}...\textgreater{}\textgreater{}) — i.e., annotation/formatting markers in the text. It fires for the bracketed token regardless of its lexical content (subword pieces like " un", " mid", " sub", short words like "cold"/"Climate", punctuation like ":"/"{]}"/"-\textgreater{}", or formatting markers like "**"), so it signals the presence of a \textless{}\textless{}...\textgreater{}\textgreater{}-wrapped token rather than a specific semantic category. \\
Markers                         & 18530      &  tokens enclosed in double angle brackets (\textless{}\textless \textgreater{}\textgreater{}) — i.e., short inline-marked words or labels (pronouns, prepositions, short nouns/labels like "Answers" or code markers) used for emphasis or markup.                                                                                                                                                                                                                                                                                                              \\
Markers                         & 7941       &  double-angle-bracket markup (\textless{}\textless and \textgreater{}\textgreater{}) used as UI/markup chevrons—typically bracketing language names or navigation/selection indicators (e.g., highlighting the selected language or menu item).                                                                                                                                                                                                                                                                                                                 \\
Markers                         & 2811       &  placeholder markers like \textless{}\textless \textgreater{}\textgreater and \textless{}\textless{}.\textgreater{}\textgreater that indicate omitted or redacted characters/tokens (often punctuation such as a dot, separators, or line breaks) in scraped text—commonly appearing inside URLs, filenames, shell commands, code snippets, and forum/blog excerpts.                                                                                                                                                                                            \\
Markers                         & 22732      &  tokens enclosed in double-angle brackets (\textless{}\textless{}...\textgreater{}\textgreater{}): annotated or highlighted short items — UI labels/call-to-action words (e.g., "Now"), single-letter coordinate directions ("S"), punctuation/arrow symbols ("-\textgreater{}"), and brief highlighted words like "people" or "social."                                                                                                                                                                                                                        \\
Markers                         & 18900      &  the token "\textless{}?\textgreater{}" — an inline/questioning punctuation marker (a question mark inside angle brackets) used to indicate uncertainty, a rhetorical/questioned remark, or an editorial query in informal prose and code-comment contexts.                                                                                                                                                                                                                                                                                                     \\
Markers                         & 16265      &  tokens that appear inside double-angle-bracket markup (\textless{}\textless \textgreater{}\textgreater{}) in scraped web text — i.e., UI/website navigation labels, section or link anchors, short structural words/punctuation used in headings, menus, and inline site navigation.                                                                                                                                                                                                                                                                           \\
Markers                         & 9756       &  placeholder/annotation tokens enclosed in double angle brackets (e.g., \textless{}\textless{}1\textgreater{}\textgreater{}, \textless{}\textless{}2\textgreater{}\textgreater{}, \textless{}\textless{}-\textgreater{}\textgreater{}, \textless{}\textless{}Cl\textgreater{}\textgreater{}, \textless{}\textless{}M\textgreater{}\textgreater{}). It fires on small functional or formatting fragments — numeric placeholders, hyphen markers, or short letter fragments used as tokenization/markup artifacts in the text.                                    \\
Markers                         & 13294      &  tokens that were highlighted/marked up in web text — typically short, common words or adjectives (e.g., a, and, short, quality, antique, thick, \&) that appear in isolation inside double angle brackets, i.e., emphasized or linked terms in scraped content.                                                                                                                                                                                                                                                                                                \\
Markers                         & 152        &  tokens that are explicitly marked/annotated with double-angle brackets (\textless{}\textless{}...\textgreater{}\textgreater{}), i.e., highlighted or labeled tokens (often section headings or target words) in the text.                                                                                                                                                                                                                                                                                                                                      \\
Markers                         & 51         &  the opening curly brace "\{" token — the start of JS expressions, object literals and style objects, and the beginning of CSS/template literal blocks (e.g., in JSX, inline styles, and styled-components).                                                                                                                                                                                                                                                                                                                                                    \\
Markers                         & 14028      &  tokens that appear inside double angle-bracket annotations (\textless{}\textless{}...\textgreater{}\textgreater{}), i.e., placeholder/markup tokens — often acronyms, abbreviations, subword fragments, or punctuation shown as examples.                                                                                                                                                                                                                                                                                                                      \\
Markers                         & 19336      &  tokens wrapped in double angle brackets — especially the placeholder "\textless{}\textless{}" . "\textgreater{}\textgreater{}" used to obfuscate periods in URLs/domain names (e.g., help.syncfusion\textless{}\textless{}.\textgreater{}\textgreater{}com, intuit\textless{}\textless{}.\textgreater{}\textgreater{}com) and, more generally, bracketed single-word annotations like \textless{}\textless{}social\textgreater{}\textgreater or \textless{}\textless{}water\textgreater{}\textgreater{}.                                                       \\
Markers                         & 13962      &  tokens enclosed in double angle‑bracket markup (\textless{}\textless{}...\textgreater{}\textgreater{}): it detects editorial/formatting annotations—often small function words or digits inside \textless{}\textless \textgreater{}\textgreater (e.g., parts of numbers like the “22” in Avogadro’s 6.022×10\textasciicircum{}23 or inserted words like \textless{}\textless{}the\textgreater{}\textgreater{}, \textless{}\textless{}of\textgreater{}\textgreater{}, \textless{}\textless{}curing\textgreater{}\textgreater{}).                                \\
Markers                         & 16758      &  tokens enclosed in double angle brackets (\textless{}\textless \textgreater{}\textgreater{}) — i.e., highlighted/target words — responding to the formatting cue (across parts of speech and semantics) rather than a specific lexical meaning.                                                                                                                                                                                                                                                                                                                \\
Markers                         & 2658       &  tokens that are enclosed in double-angle brackets (\textless{}\textless \textgreater{}\textgreater{}) — i.e., markup/annotation or highlighted tokens in the text — regardless of the token’s lexical class or meaning.                                                                                                                                                                                                                                                                                                                                        \\
Section boundary                & 3228       &  list-item/outline markers and formatting tokens (numbers, bullets, punctuation used to start or separate list entries) often appearing in medical or clinical explanatory text.                                                                                                                                                                                                                                                                                                                                                                                \\
Section boundary                & 26417      &  list or section-boundary markers — tokens that introduce or separate enumerated items (colons, bullets, newlines/paragraph breaks, sentence delimiters) — especially when starting lists of risk factors, symptoms, or steps in medical/health texts.                                                                                                                                                                                                                                                                                                          \\
Punctuations                    & 15331      &  sentence-final punctuation — the period/full stop (end-of-sentence markers such as “.” or “.)”), i.e., tokens that mark sentence boundaries.                                                                                                                                                                                                                                                                                                                                                                                                                   \\
Punctuations                    & 22999      &  sentence-boundary tokens — i.e., end-of-sentence punctuation or break markers (periods, dashes/other break tokens) that signal the transition from one sentence or clause to the next.                                                                                                                                                                                                                                                                                                                                                                         \\
Punctuations                    & 11923      &  end-of-sentence/paragraph boundary tokens — punctuation or break markers (e.g., periods, line breaks, ellipses, dashes) that signal the end of one sentence/section and the start of the next (often followed by a capitalized word).                                                                                                                                                                                                                                                                                                                          \\
Punctuations                    & 12378      &  sentence-ending full stops (end-of-sentence punctuation). It fires on periods and on tokens tied to those stops (e.g., the digit or next word tokenized after a period or decimal), signaling sentence boundaries.                                                                                                                                                                                                                                                                                                                                             \\
Punctuations                    & 25989      &  text boundary/transition tokens — punctuation and formatting markers (commas, periods, parentheses, dashes, section breaks) and short function words (articles, conjunctions) that signal clause or sentence boundaries or transitions.                                                                                                                                                                                                                                                                                                                        \\
Punctuations                    & 4033       &  sentence and phrase boundaries — it responds to punctuation (periods) and common space-prefixed small tokens/prefixes (e.g., " .", " of", " un", " Not", " stops") that mark ends of sentences or transitions between clauses.                                                                                                                                                                                                                                                                                                                                 \\
Punctuations                    & 16562      &  sentence boundaries — it activates on sentence-final punctuation (periods, sometimes closing parens or line breaks) that mark the end of a sentence or paragraph (often followed by the start of the next sentence).                                                                                                                                                                                                                                                                                                                                           \\
Punctuations                    & 13631      &  sentence-final periods (full stops) — i.e., end-of-sentence punctuation marking the boundary before the next sentence.                                                                                                                                                                                                                                                                                                                                                                                                                                         \\
Punctuations                    & 4451       &  sentence-final periods — the end-of-sentence (declarative) punctuation token (and more generally signals tokens occurring at sentence boundaries).                                                                                                                                                                                                                                                                                                                                                                                                             \\
Punctuations                    & 1274       &  sentence boundaries — end-of-sentence punctuation and breaks (periods, paragraph/newline breaks, bullets) that mark the end of a sentence and the start of the next unit.                                                                                                                                                                                                                                                                                                                                                                                      \\
Punctuations                    & 12869      &  textual boundary/transition markers — it fires on formatting and punctuation that signal discourse breaks (paragraph/line breaks, quotes, dashes, periods) and, more broadly, on tokens that mark sentence or clause transitions.                                                                                                                                                                                                                                                                                                                              \\
Punctuations                    & 28223      &  sentence-boundary punctuation — tokens like periods, colons (often after speaker names) and opening quotation marks — that mark the end of one sentence and the start of a new sentence or quoted/dialogue utterance.                                                                                                                                                                                                                                                                                                                                          \\
Punctuations                    & 3618       &  the sentence-final full stop (period) — the punctuation token that marks the end of a sentence/sentence boundary.                                                                                                                                                                                                                                                                                                                                                                                                                                              \\
Punctuations                    & 14384      &  sentence- or phrase-ending punctuation and closing separators — e.g., periods (including period after a closing quote), dashes/en‑dashes used as clause separators, and similar tokens that mark sentence boundaries or breaks in running text.                                                                                                                                                                                                                                                                                                                \\
Punctuations                    & 8154       &  sentence-boundary punctuation and discourse-transition markers (e.g., periods, colons, other end-of-sentence tokens) often used to mark the end of a sentence or a shift between clauses/sections in academic and historical prose.                                                                                                                                                                                                                                                                                                                            \\
Punctuations                    & 22970      &  sentence/clause boundaries and transition tokens — it lights up on end-of-sentence punctuation (periods/newlines) and on small function words that signal clause transitions (e.g., "to", "even"), often at sentence breaks in instructional/medical text.                                                                                                                                                                                                                                                                                                     \\
Punctuations                    & 27755      & s end-of-sentence punctuation (period/ellipsis) — the sentence boundary token, especially when it precedes a discourse transition or explanatory continuation (e.g., "This is because," "In addition," "Though").                                                                                                                                                                                                                                                                                                                                                \\* \bottomrule
\label{tab:feature-ids-explanations}
\end{longtable}

\end{document}